%% file: LREC2026/lrec2026-example.tex
\documentclass[10pt, a4paper]{article}

\usepackage[final]{LREC2026/lrec2026} 
\usepackage{booktabs} 
\usepackage{amsmath}
\usepackage{amssymb}
\usepackage{algpseudocode}
\usepackage{algorithm}
\usepackage{authblk}
\usepackage[most]{tcolorbox}
\usepackage{tcolorbox}
\tcbuselibrary{listings,breakable}

\usepackage{listings}

\lstdefinestyle{promptstyle}{
  basicstyle=\ttfamily\footnotesize,
  breaklines=true,
  columns=fullflexible,
  keepspaces=true,
  showstringspaces=false
}

\newtcblisting{promptbox}[2][]{
  colback=gray!5,
  colframe=gray!60,
  boxrule=0.4pt,
  arc=2pt,
  left=4pt,
  right=4pt,
  top=4pt,
  bottom=4pt,
  listing only,
  listing options={style=promptstyle},
  breakable,
  title=#2,#1
}

\title{sebis at CRF Filling 2026: A Two-Stage Local LLM Pipeline for Medical CRF Filling}

\name{Katharina Sommer$^{\dagger}$, Tristan Till$^{\dagger}$, Florian Matthes} 

\address{Technical University of Munich, TUM School of Computation, Information and Technology\\
         \{kathi.sommer, tristan.till, matthes\}@tum.de\\}

\abstract{
The extraction of structured clinical information from unstructured EHR notes is a persistent bottleneck in healthcare informatics. While large language models (LLMs) offer high performance, their deployment in clinical settings is hindered by privacy risks, inference costs, and the tendency to hallucinate beyond textual evidence. We address these challenges for the CL4Health 2026 Case Report Form (CRF) filling task by proposing a fully local, domain-adapted pipeline using the MedGemma-27B model. Our two-stage architecture, which separates binary presence classification from value extraction, enforces strict adherence to textual evidence and ensures deterministic outputs for negated, uncertain, or unknown states. By leveraging item-specific, few-shot in-context learning without external API calls or fine-tuning, our approach achieves a macro-F1 score of 0.55 on the official English test track. This result secures second place among all locally-hosted, open-source submissions. Our work demonstrates that privacy-preserving, on-premise LLM pipelines can achieve near-competitive performance with proprietary frontier models, providing a practical, data-sovereign framework for clinical NLP.
 \\ \newline  }

\begin{document}

\maketitleabstract
\renewcommand{\thefootnote}{\fnsymbol{footnote}}
\footnotetext[2]{These authors contributed equally to this work.}
\renewcommand{\thefootnote}{\arabic{footnote}}

\section{Introduction}
The extraction of structured clinical information from unstructured free-text notes remains a persistent bottleneck in healthcare informatics. Electronic health records (EHRs) contain rich patient narratives generated in high-volume settings such as emergency departments, yet converting these narratives into standardized, research-grade formats requires laborious manual review by clinicians or trained abstractors. Case Report Forms (CRFs) exemplify this challenge: they are predefined, study-specific instruments that demand consistent encoding of dozens of variables (e.g., comorbidities, vital signs, laboratory values, symptom chronicity) for each patient. The CRF Filling shared task \cite{ferrazzi-etal-2026-crf-filling} at the CL4Health 2026 workshop directly addresses this pain point by asking participants to automatically populate a CRF for dyspnea patients seen in emergency departments, using only clinical notes written in English or Italian. Success in the task would accelerate observational research, reduce documentation burden, and enable real-time, patient-friendly summarization.

While traditional rule-based and supervised clinical information extraction systems achieve high precision in narrow domains, they lack the scalability and zero-shot flexibility required for diverse clinical settings. Large language models accessed via commercial APIs offer impressive flexibility yet introduce critical drawbacks in real-world deployment, such as privacy and regulatory risks when sensitive patient data leave institutional firewalls, prohibitive inference costs and latency at scale, and uncontrolled hallucinations or assumption-making that violate the strict \textit{“do not infer beyond the text”} requirement of CRF filling.

Our submission closes these gaps through a fully local, domain-adapted LLM pipeline that requires only the official training set for few-shot guidance. By leveraging the open-weight MedGemma-27B model \cite{sellergren2025medgemma}, we guarantee data sovereignty and HIPAA/GDPR compliance while maintaining clinical-domain knowledge that generic models lack. The two-stage architecture, presence classification followed by value extraction, combined with engineered prompts and per-item few-shot exemplars, enforces strict adherence to the text and eliminates extraneous output. This design is both novel for the shared task and practically deployable in hospital environments. In summary, our core contributions are: (1) an applied demonstration of using a locally hosted medical LLM for CRF filling under strict privacy constraints; (2) a reusable, classification-then-extraction prompting framework that handles binary presence detection, categorical selection, and continuous “measured” values within a single inference pass; and (3) an efficient, zero-training implementation that achieves competitive performance in the English track while respecting the task’s emphasis on minimal supervision.

\begin{figure*}[t]
\centering
\includegraphics[width=0.95\textwidth]{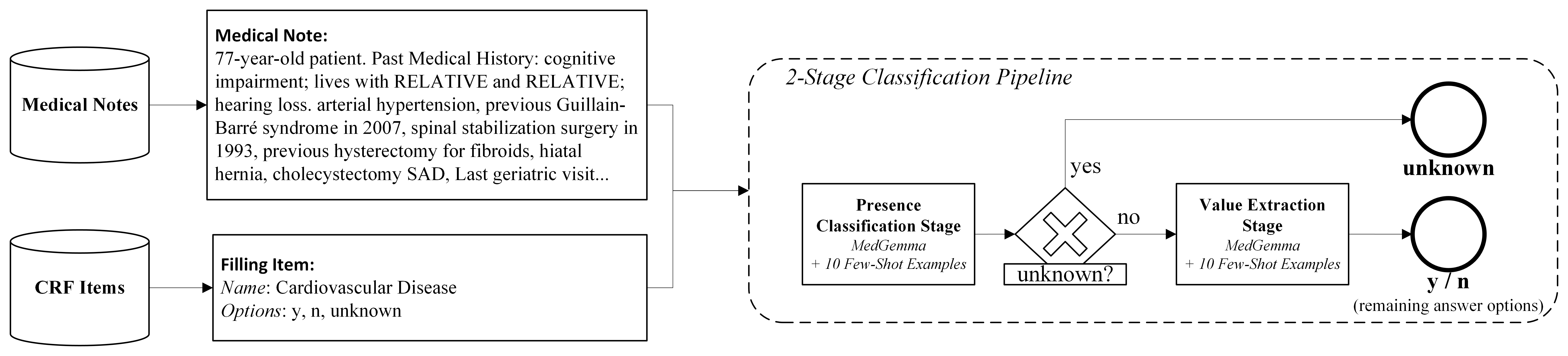}
\caption{Overview of the two-stage classification-then-extraction pipeline. For each clinical note and CRF item, the system first determines presence via a classification prompt. If the presence is confirmed, the system proceeds to the Value Extraction Stage to output the specific categorical or measured value.}
\label{fig:pipeline}
\end{figure*}

\begin{figure}[]
\centering
\includegraphics[width=0.45\textwidth]{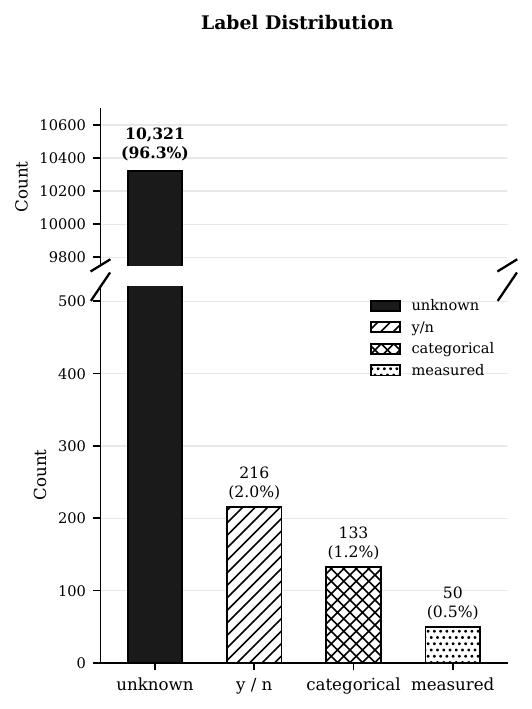}
\caption{Label distribution of the training set across the four item types. The dataset exhibits a severe class imbalance, with \textit{"unknown"} representing the vast majority of labels, emphasizing the challenge of minimizing false positives during extraction.}
\label{fig:label-distribution}
\end{figure}

\section{Related Work}
Large Language Models (LLMs) have established few-shot prompting as a powerful paradigm for clinical information extraction under low-resource constraints. \citet{agrawal2022large} demonstrated that models such as InstructGPT can achieve strong zero- and few-shot performance on clinical entity and relation extraction tasks directly from unstructured notes, relying solely on in-context examples without any fine-tuning. This approach aligns closely with the CL4health CRF-filling requirements, where systems must respect strict textual evidence, handle negations and uncertainty, and output deterministic values across multiple predefined items.

Further research has addressed the critical need for privacy-preserving deployment, focusing on fully local, open-source LLMs. \citet{wiest2024llm} introduced LLM-AIx, an on-premise pipeline based on Llama 2 that extracts structured information from clinical free text while keeping all patient data within institutional infrastructure. This line of work was extended by \citet{builtjes2025leveraging}, who showed that open-source generative LLMs deliver viable zero-shot performance on clinical extraction benchmarks in resource-constrained and multilingual settings, and by \citet{richter2025medication}, who achieved new state-of-the-art results on end-to-end medication information extraction through lightweight fine-tuning of local models on both English and German corpora.

Domain-specialized medical LLMs have further bridged the gap between general-purpose models and clinical precision. MedGemma-27B \cite{sellergren2025medgemma}, specialized on medical text, has established leading performance on clinical reasoning and extraction benchmarks while remaining fully open-weight and locally deployable. Our two-stage classification-then-extraction pipeline leverages exactly this model in a quantized, on-premise setup, synthesizing the few-shot flexibility shown by \citep{agrawal2022large}, the privacy guarantees of local open-source pipelines, and MedGemma's domain knowledge into a practical solution for the CRF-filling shared task.

\begin{figure*}[t]
\centering
\includegraphics[width=0.95\textwidth]{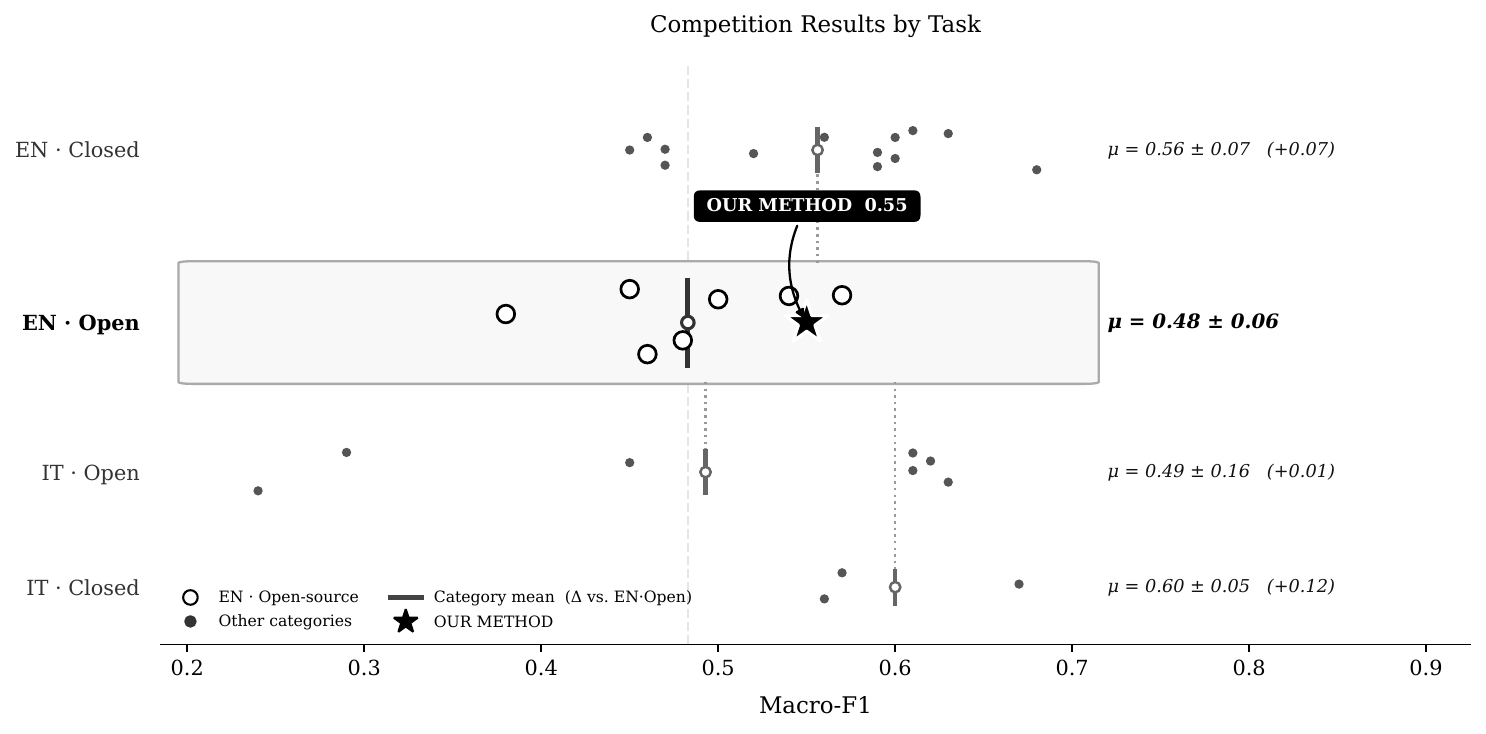}
\caption{Diagram of the CL4Health 2026 CRF-filling leaderboard, split into detection language and open- vs. closed-weights models. On average, results on the Italian test set yield superior results to the English subset. Similarly, the use of closed-weights models shows higher Macro-F1 scores on average. Our method ranks \#2 for submissions in English with open-weights models and \#17 overall.}
\label{fig:leaderboard}
\end{figure*}

\section{Methodology}
\subsection{Data}
We use the official English-track datasets released for the shared task \cite{kaczmarek2026automaticfillingcasereport, ferrazzi-etal-2025-converting}. The training split contains clinical notes together with gold-standard annotations structured as a list of dictionaries. The test set provides the raw clinical notes and document identifiers required for submission. No synthetic data or external corpora were added, preserving the low-resource spirit of the task. Note that Figure~\ref{fig:label-distribution} shows a severe class imbalance on a per-sample, per-CRF-item basis, in which \textit{"unknown"} dominates.

\subsection{Model}
Inference is performed with instruction-tuned MedGemma-27B (GGUF Q8\_K\_XL quantization) with temperature set to 0.

\subsection{Pipeline Architecture}
The system implements a two-stage, per-document, per-item loop, as shown in Figure \ref{fig:pipeline}. The algorithm for this approach can be seen in Algorithm  \ref{alg:llm_pipeline}.

\paragraph{Presence Classification Stage:}We construct 10 few-shot examples for each category from the training. As the training set consists of only 10 samples, we use each sample to construct a few-shot example for every category. Then, the ground-truth labels are binarized to indicate whether the ground-truth is known (i.e., the answer is \textit{"unknown"}) or an actual value. For \textit{"unknown"}, we set the binary label to \texttt{n} and for all other cases to \texttt{y}. A classification prompt instructs the model to return exactly one character (\texttt{y} / \texttt{n}) and includes explicit instructions on negations and uncertainty. If the model response is \texttt{y}, we proceed with the Value Extraction Stage. Otherwise, \textit{"unknown"} is recorded and the loop is aborted.

\paragraph{Value Extraction Stage:}We construct another 10 few-shot examples using the actual ground-truth values, which now include the full option and or \textit{"measured"} string. If the target item's option list begins with \textit{"measured"}, we invoke a special measured-value prompt to task the model with extracting the specific value. Otherwise, a categorical prompt list is invoked, listing all and enforcing one of the allowed options.

\begin{algorithm}[t]
\caption{Two-Step LLM Extraction Pipeline (per document)}
\label{alg:llm_pipeline}
\begin{algorithmic}[1]

\Require Clinical note $T$, option table $O$
\Ensure Predictions $P$

\State $P \gets \emptyset$

\For{each row $r \in O$}

    \State $i \gets r.\text{item}$
    \State $\textit{opt} \gets r.\text{options}$

    \Comment{Step 1: Presence classification}

    \State $\textit{fs}_{c} \gets \text{create\_few\_shot}(i)$
    \State $\textit{p}_{c} \gets \text{create\_prompt\_classification}(i, T, \textit{fs}_{c})$
    \State $\hat{y} \gets \text{LLM}(\textit{p}_{c})$

    \If{$\hat{y} = \text{`y'}$}

        \Comment{Step 2: Value extraction}

        \State $\textit{fs}_{v} \gets \text{create\_few\_shot}(i, \textit{opt})$

        \If{$\textit{opt} = \text{``measured''}$}
            \State $\textit{p} \gets \text{prompt\_measured}(i, T, \textit{fs}_{v})$
        \Else
            \State $\textit{p} \gets \text{prompt\_answer}(i, \textit{opt}, T, \textit{fs}_{v})$
        \EndIf

        \State $\hat{v} \gets \text{LLM}(\textit{p})$

    \Else
        \State $\hat{v} \gets \text{``unknown''}$
    \EndIf

    \State $P \gets P \cup \{(i, \hat{v})\}$

\EndFor

\State \Return $P$

\end{algorithmic}
\end{algorithm}



\begin{table*}[t]
    \centering
    \begin{tabular}{cccccccc}
    \toprule
        MedGemma & Few-Shot & Two-Step & Macro-F1 ($\uparrow$) & \%-gain ($\uparrow$) & TPR ($\uparrow$) & FPR ($\downarrow$) & FNR ($\downarrow$) \\
    \midrule
         \checkmark & X & X & 0.330& + 0.00\% & 7.35\% & 92.35\% & \textbf{0.30\%} \\
         \midrule
         \checkmark & \checkmark & X & \underline{0.572} & \underline{+73.33\%} & \underline{40.00\%} & \underline{50.46\%} & 9.54\% \\
         \checkmark & X & \checkmark & 0.423 & +28.18\% & 26.15\% & 66.00\% & \underline{7.85\%} \\
         \checkmark & \checkmark & \checkmark & \textbf{0.604} & \textbf{+83.03\%} & \textbf{40.56\%} & \textbf{44.66\%} & 14.79\% \\
         \bottomrule
    \end{tabular}
    \caption{Ablations on the internal development set quantifying the impact of the two-stage architecture and few-shot prompting on performance metrics (Macro-F1, TPR, FPR, and FNR). The baseline is a zero-shot MedGemma-27B model. Results demonstrate the additive benefits of the two-stage gatekeeping mechanism, reducing false-positive rates and boosting both recall and overall F1.}
    \label{tab:abaltion}
\end{table*}

\section{Results}
\subsection{Leaderboard Results}
Our final submission achieves a macro-F1 score of 0.55 on the official hidden test set of the CL4Health 2026 CRF-filling shared task. As visualized in Figure~\ref{fig:leaderboard}, this result places us 17th out of 32 total submissions worldwide. Restricting the view to the English-only track, we rank 11th out of 22 participants. Most importantly, among the seven fully local and open-source systems submitted to the English track, our MedGemma-based pipeline secures a strong 2nd place, coming within 0.02 macro-F1 of the top local open-source entry (0.57) while remaining more than 0.13 points behind the overall winner (a closed-source English system at 0.68). These rankings highlight that our privacy-preserving, on-premise approach delivers highly competitive performance without relying on external APIs or proprietary models.

\subsection{Development Phase Ablations}

To justify each design decision, we performed systematic ablations on an internal development split derived from the official training set. Starting from a naive zero-shot MedGemma-27B baseline (macro-F1 = 0.330), simply adding the two-stage classification-then-extraction architecture increased performance to 0.423, confirming that first verifying presence before attempting value extraction substantially reduces false-positive predictions on absent items. Incorporating item-specific 10-shot examples without the two-stage design further raised the score to 0.572, demonstrating the critical value of in-context learning. Finally, the complete pipeline including two-stage prompting plus few-shot examples reached 0.604 macro-F1 (+0.032 over the few-shot-only variant), proving that all components act additively. These gains directly motivated our final architecture and explain why the same pipeline translated into the strong leaderboard ranking reported above.
\section{Discussion}

The leaderboard results (Figure~\ref{fig:leaderboard}) confirm a performance gap between closed-weight and open-weight models. While proprietary frontier models maintain advantages in scale and alignment, our second-place finish among local, open-source submissions demonstrates that domain-adapted models like MedGemma-27B can achieve near-competitive performance while maintaining data sovereignty: a requirement for clinical deployment.

Ablation studies (Table \ref{tab:abaltion}) quantify the additive benefits of our pipeline. A naive zero-shot baseline is overly conservative, missing most positive cases. Integrating few-shot examples provides the single largest gain, teaching the model to identify positive evidence. Separately, the two-stage gatekeeper improves precision by reducing spurious extractions. Combining both components yields the optimal error profile, confirming that few-shot guidance teaches detection while the classification stage acts as a filter against over-prediction.

Our primary development challenge was mitigating false positives, particularly the model’s tendency to over-interpret negated or distant mentions as positive evidence. Reliably distinguishing between a ruled-out label and missing information remains a known difficulty in clinical NLP. We addressed this through explicit negation/uncertainty instructions, the two-stage gatekeeper, and strict output constraints. Despite these improvements, residual false positives remain the primary bottleneck. Future work will explore lightweight preference-based alignment, retrieval-augmented few-shot selection, and ensemble-based uncertainty estimation to further suppress over-confident extractions without sacrificing recall.

\section{Future Work}
While our two-stage pipeline provides a privacy-preserving and efficient alternative to commercial APIs, several areas of future work remain. First, future work will explore automated retriever-based selection to dynamically choose the most relevant exemplars for each patient note. 

Second, although our presence-classification gatekeeper reduces false positives, the model still occasionally misinterprets negated mentions as positive evidence. Future developments will explore lightweight, parameter-efficient fine-tuning (e.g., LoRA) on domain-specific negation-handling datasets to improve precision. Finally, our pipeline currently operates per-item; integrating a cross-item consistency checker could further reduce internal contradictions (e.g., ensuring mutually exclusive clinical symptoms are not both marked as present).

\section{Conclusion}

In this paper, we presented a privacy-preserving, two-stage pipeline for CRF filling using the MedGemma-27B model. By combining a presence-classification gatekeeper with value-extraction prompts, our system effectively mitigates the common LLM pitfalls of over-interpretation and hallucination. Our results confirm that local, open-weight models, when paired with strategic few-shot guidance, can bridge the performance gap between institutional, on-premise solutions and closed-source commercial APIs. This approach provides a scalable and secure alternative for clinical extraction tasks where data sovereignty is a non-negotiable requirement. Future work will focus on integrating preference-based alignment and ensemble-based uncertainty estimation to further refine the precision of clinical extractions in high-stakes healthcare environments.

\section{Limitations}
Experiments were conducted only on English data, limiting the scope to English-only health records and omitting comparisons of how the approach and model perform in non-English languages.

Furthermore, the dependency on few-shot in-context learning makes the system sensitive to the quality and diversity of training examples. Without at least a few gold standard examples, the accuracy of this approach cannot be guaranteed. 

Lastly, the original dataset is in Italian, the dataset used for experiments is translated into English by the dataset authors. Therefore, translation artifacts could be present in the dataset, potentially skewing the results of the task. 

\section{Ethical Statement}
This work uses anonymised patient data collected at San Giovanni Bosco Hospital. The data was anonymised by the dataset authors and made publicly available. Therefore, no ethical considerations apply.

\section{Bibliographical References}
\bibliographystyle{LREC2026/lrec2026-natbib}
\bibliography{LREC2026/lrec2026-example}
\label{lr:ref}
\bibliographystylelanguageresource{LREC2026/lrec2026-natbib}
\bibliographylanguageresource{LREC2026/languageresource}

\include{LREC2026/appendix}
\end{document}

%% file: LREC2026/appendix.tex
\appendix
\section{Appendix}
\subsection{Prompts}
The following prompt shows the first step of the pipeline, which classifies whether a given attribute appears in the clinical note. The model is expected to return either \texttt{y} or \texttt{n} depending on whether the given category is present in the clinical text or not. 
\begin{promptbox}[title=Presence Classification Prompt (1$^{st}$ Step)]
aYou are a clinical assistant helping a doctor extract information from clinical text.

For a given category, determine whether information about it is present in the text.

Instructions:
    - Only return a single character: 'y' for Yes (present) or 'n' for No (absent).
    - Do NOT make any assumptions beyond the text.
    - Pay attention to negations, unknown, or unclear mentions.
    - Do not add explanations or extra text.

Few-shot examples:
{few_shot}

Now check the following category in the given clinical note:

Category: {category}

Clinical Text: {text}

Answer:
\end{promptbox}

The next two prompts show the prompts for the actual CRF task. There are two types of categories: one is selecting the correct value from a list of options (see the first prompt), and the other is extracting an exact numerical value with a unit from the text (see the second prompt). The model is prompted with either one of the options depending on the category.
\begin{promptbox}[title=Value Extraction Prompt (2$^{nd}$ Step)]
aYou are a clinical assistant helping a doctor extract information from clinical text. 
Your task is to select the correct value for a given category based ONLY on the text. 
You MUST choose from the provided options and nothing else.

Instructions:
    - Only return ONE of the provided options.
    - The option 'unknown' means the clinical text does NOT provide enough information about this category.
    - Do NOT make assumptions beyond the text.
    - Pay attention to negations, uncertainty, or unclear mentions.
    - Do not provide explanations or extra text; return only the single selected option.

Few-shot examples: {few_shot}

Now evaluate the following:

Category: {category}
Options: {option}
Clinical Text: {text}

Answer:
\end{promptbox}

\begin{promptbox}[title=Value Extraction Measured Prompt (2$^{nd}$ Step)]
sYou are a clinical assistant helping a doctor extract information from clinical text. 
Your task is to find the value of a given category from the text. 
You MUST choose from the text itself and include the unit if present.

Instructions:
    - Extract the exact value from the clinical text, including units if available.
    - If the information is not mentioned in the text, return 'unknown'.
    - Do NOT make assumptions or invent values beyond what is in the text.
    - Pay attention to negations, uncertainty, or unclear mentions.
    - Do not provide explanations or extra text; return only the single extracted value or 'unknown'.

Few-shot examples: {few_shot}

Now evaluate the following:

Category: {category}
Clinical Text: {text}

Answer:
\end{promptbox}

The following prompt shows how one few-shot example looks like for the classification task in the 1$^{st}$ Step. The answer is either \texttt{y} or \texttt{n}, depending on whether the category appears in the clinical text or not.
\begin{promptbox}[title=Presence Classification Few Shot Template (1$^{st}$ Step)]
a------------------------------------
Category: {category}
Clinical Text: {note}
Answer: {value}
------------------------------------
\end{promptbox}

The next prompt shows one few-shot example for the actual CRF task. The answer is either one of the given options or the exact extracted value with a unit. 
\begin{promptbox}[title=Value Extraction Few Shot Template (2$^{nd}$ Step)]
a------------------------------------
Category: {category}
Options: {options}
Clinical Text: {note}
Answer: {value}
------------------------------------
\end{promptbox}

\subsection{Prompt Examples}
This section gives an example prompt for every stage of the pipeline. Assume we want to fill in the category "chronic pulmonary disease" which has the following options: ['certainly chronic' 'possibly chronic' 'certainly not chronic' 'unknown']. The following examples shows how the prompt would look like. For better readability only 2 few-shot samples are shown.

\begin{promptbox}[title=1$^{st}$ Step - Presence Classification]
aYou are a clinical assistant helping a doctor extract information from clinical text.

For a given category, determine whether information about it is present in the text.

Instructions:
    - Only return a single character: 'y' for Yes (present) or 'n' for No (absent).
    - Do NOT make any assumptions beyond the text.
    - Pay attention to negations, unknown, or unclear mentions.
    - Do not add explanations or extra text.

Few-shot examples:
    Category: chronic pulmonary disease
    Clinical Text: TRIAGE:
    Reports having fallen accidentally, impacting:
        - occipital region with no loss of consciousness or concussion
        - right trochanter
        - right knee
        - right ankle
    CS 15_Cincinnati negative_fluent speech_isochoric, isocyclic, and photoreactive pupils. No rigidity.
    Conscious, lucid, oriented, not agitated.
    wearing rigid cervical collar
    BP 120/80 mm/hg, SpO2 99
        
    MEDICAL ASSESSMENT:
        reports accidental fall to the ground at home following loss of support on right leg (spontaneous femur fracture?). Impact to the right side of the body, as described in triage. Currently experiencing pain in the right femur and right elbow. Head trauma without loss of consciousness. No vomiting after the episode.
        No other details at the moment
        
    Past Medical History:
        - Arterial hypertension
        - CAD (reports quadruple BPAC)
        - Diabetes mellitus type 2
        - Left hip prosthesis
        
    Therapy: metformin 500 mg, iperten 1/2 tablet, enalapril 5 mg, omnic 0.4 mg, seloken 100 mg 1/2 tablet x 2, cardioASA, torvast 20 mg, contramala 15 drops, tachipirina AD, omnepraozlo 20 mg
    
    ALLERGIES: none known

    Answer: n

------------------------------------

    Category: chronic pulmonary disease
    Clinical Text: Alert, oriented, cooperative
    BP 150/80 mmHg HR 78r bpm SpO2 96
    Pulmonary examination: Bilateral diffuse vesicular murmur, no additional sounds.
    Cardiac examination: valid, rhythmic tones
    Abdominale examination: distended, palpable, not tender or painful, Murphy and Blumberg negative, peristalsis present
    Lower limbs: no peripheral edema; peripheral pulses hyposphygmic, bilaterally palpable up to the popliteal artery, with no pedidium pulse palpable bilaterally; dry necrosis left first toe, mild cyanosis in the remaining toes of the left foot; no sensory or motor deficits in the left lower limbs
    ECG unchanged from the previous one

    Answer: n
------------------------------------
{8 more few-shot samples - skipped here for readability}
------------------------------------

Now check the following category in the given clinical note:

Category: chronic pulmonary disease

Clinical Text: 77-year-old patient.
Past Medical History: cognitive impairment; lives with RELATIVE and RELATIVE; hearing loss. arterial hypertension, previous Guillain-Barre syndrome in 2007
spinal stabilization surgery in 1993, previous hysterectomy for fibroids, hiatal hernia, cholecystectomy
SAD, Last geriatric visit PHONE_NUMBER: MRI cerebral vasculopathy - presence of buccal dyskinesias - slow walking with risk of falls
Therapy: pantoprazole 40-olmesartan 1/2- xanax 0.25- contramal 4 drops x2, levopraid 1/2 x 2, folic acid, stilnox,
RELATIVE PHONE_NUMBER

Recent Medical History: patient arrived last night for an episode of sudden illness characteized by a scream followed by generalized TC tremors in all 4 limbs; this morning a second fit was seen in the Emergency and Acceptance Department, TC in all 4 limbs, preceded by a scream.
Performed in the Emergency Department: blood tests: no abnormalities (normal electrolytes), no fever; experiencing sleep loss due to frequent nighttime awakenings. Also performed brain CT brain without contrast which showed a picture of discrete diffuse cortico-subcortical atrophy, no acute manifestations; known chronic vascular disease.

Neurological Examination: patient alert, cooperative, rather quiet, partially oriented in time/place; no clear oculomotor deficits, blinks at bilateral visual threat, no underleveling at Mingazzini I; raises lower limbs alternately for 4-5 sec with symmetrical controlled descent. No appreciable sensory deficits. Bilateral plantar reflex in flexion.

Conclusions: generalized TC seizures (2 episodes) due to moderate cerebral atrophy and cognitive impairment.
Recommendation to start gardenale 50 mg 1 tablet in the evening. Clinical observation until tomorrow morning; if there are no other episodes or other illnesses, discharge tomorrow with indication to perform  EEG with priority B and check-up in 6 months at epilepsy clinic with blood phenobarbital levels, a complete blood count, and liver and kidney function tests..

        Answer:
\end{promptbox}

Assuming that the answer of the LLM to the above prompt is \texttt{y}, and because the option is not "measured" but categorical, the following prompt is sent to the LLM. Again, only 2 of the few-shot prompts are shown.

\begin{promptbox}[title=2$^{nd}$ Step - Value Extraction]
You are a clinical assistant helping a doctor extract information from clinical text. 
Your task is to select the correct value for a given category based ONLY on the text. 
You MUST choose from the provided options and nothing else.

Instructions:
    - Only return ONE of the provided options.
    - The option 'unknown' means the clinical text does NOT provide enough information about this category.
    - Do NOT make assumptions beyond the text.
    - Pay attention to negations, uncertainty, or unclear mentions.
    - Do not provide explanations or extra text; return only the single selected option.

Few-shot examples:
    Category: chronic pulmonary disease
    Options: ['certainly chronic' 'possibly chronic' 'certainly not chronic' 'unknown']
    Clinical Text: TRIAGE: 
    Reports having fallen accidentally, impacting:
    - occipital region with no loss of consciousness or concussion
    - right trochanter 
    - right knee
    - right ankle
    CS 15_Cincinnati negative_fluent speech_isochoric, isocyclic, and photoreactive pupils. No rigidity.
    Conscious, lucid, oriented, not agitated.
    wearing rigid cervical collar
    BP 120/80 mm/hg, SpO2 99
    
    MEDICAL ASSESSMENT: 
    reports accidental fall to the ground at home following loss of support on right leg (spontaneous femur fracture?). Impact to the right side of the body, as described in triage. Currently experiencing pain in the right femur and right elbow. Head trauma without loss of consciousness. No vomiting after the episode. 
    No other details at the moment
    
    Past Medical History: 
    - Arterial hypertension
    - CAD (reports quadruple BPAC)
    - Diabetes mellitus type 2
    - Left hip prosthesis
    
    Therapy: metformin 500 mg, iperten 1/2 tablet, enalapril 5 mg, omnic 0.4 mg, seloken 100 mg 1/2 tablet x 2, cardioASA, torvast 20 mg, contramala 15 drops, tachipirina AD, omnepraozlo 20 mg
    ALLERGIES: none known

    Answer: unknown
    
------------------------------------
    
    Category: chronic pulmonary disease
    Options: ['certainly chronic' 'possibly chronic' 'certainly not chronic' 'unknown']
    Clinical Text: Alert, oriented, cooperative
    BP 150/80 mmHg HR 78r bpm SpO2 96
    Pulmonary examination: Bilateral diffuse vesicular murmur, no additional sounds.
    Cardiac examination: valid, rhythmic tones
    Abdominale examination: distended, palpable, not tender or painful, Murphy and Blumberg negative, peristalsis present
    Lower limbs: no peripheral edema; peripheral pulses hyposphygmic, bilaterally palpable up to the popliteal artery, with no pedidium pulse palpable bilaterally; dry necrosis left first toe, mild cyanosis in the remaining toes of the left foot; no sensory or motor deficits in the left lower limbs
    ECG unchanged from the previous one

    Answer: unknown
    
------------------------------------
{8 more few-shot samples - skipped here for readability}
------------------------------------

Now evaluate the following:

Category: chronic pulmonary disease
Options: ['certainly chronic' 'possibly chronic' 'certainly not chronic' 'unknown']
        Clinical Text: 77-year-old patient.
Past Medical History: cognitive impairment; lives with RELATIVE and RELATIVE; hearing loss. arterial hypertension, previous Guillain-Barre syndrome in 2007
spinal stabilization surgery in 1993, previous hysterectomy for fibroids, hiatal hernia, cholecystectomy
SAD, Last geriatric visit PHONE_NUMBER: MRI cerebral vasculopathy - presence of buccal dyskinesias - slow walking with risk of falls
Therapy: pantoprazole 40-olmesartan 1/2- xanax 0.25- contramal 4 drops x2, levopraid 1/2 x 2, folic acid, stilnox,
RELATIVE PHONE_NUMBER

Recent Medical History: patient arrived last night for an episode of sudden illness characteized by a scream followed by generalized TC tremors in all 4 limbs; this morning a second fit was seen in the Emergency and Acceptance Department, TC in all 4 limbs, preceded by a scream.
Performed in the Emergency Department: blood tests: no abnormalities (normal electrolytes), no fever; experiencing sleep loss due to frequent nighttime awakenings. Also performed brain CT brain without contrast which showed a picture of discrete diffuse cortico-subcortical atrophy, no acute manifestations; known chronic vascular disease.

Neurological Examination: patient alert, cooperative, rather quiet, partially oriented in time/place; no clear oculomotor deficits, blinks at bilateral visual threat, no underleveling at Mingazzini I; raises lower limbs alternately for 4-5 sec with symmetrical controlled descent. No appreciable sensory deficits. Bilateral plantar reflex in flexion.

Conclusions: generalized TC seizures (2 episodes) due to moderate cerebral atrophy and cognitive impairment.
Recommendation to start gardenale 50 mg 1 tablet in the evening. Clinical observation until tomorrow morning; if there are no other episodes or other illnesses, discharge tomorrow with indication to perform  EEG with priority B and check-up in 6 months at epilepsy clinic with blood phenobarbital levels, a complete blood count, and liver and kidney function tests..

Answer:
\end{promptbox}